\documentclass[11pt,letterpaper]{article}
\usepackage{emnlp2017}
\usepackage{times}
\usepackage{latexsym}
\usepackage{amsmath}
\usepackage{amsfonts}
\usepackage{graphicx}
\usepackage{booktabs}
\usepackage{multirow}
\usepackage[linesnumbered,ruled,vlined]{algorithm2e}
\usepackage[font={small}]{caption}

\emnlpfinalcopy

\def\emnlppaperid{1047}

\title{DeepPath: A Reinforcement Learning Method for\\ Knowledge Graph Reasoning}

\author{Wenhan Xiong \and Thien Hoang \and  William Yang Wang\\
   Department of Computer Science\\
   University of California, Santa Barbara\\
   Santa Barbara, CA 93106 USA\\
  {\tt \{xwhan,william\}@cs.ucsb.edu, thienhoang@umail.ucsb.edu}}

\date{}

\begin{document}
\maketitle
\begin{abstract}
We study the problem of learning to reason in large scale knowledge graphs (KGs). More specifically, we describe a novel reinforcement learning framework for learning multi-hop relational paths: we use a policy-based agent with continuous states based on knowledge graph embeddings, which reasons in a KG vector space by sampling the most promising relation to extend its path. In contrast to prior work, our approach includes a reward function that takes the \textbf{accuracy}, \textbf{diversity}, and \textbf{efficiency} into consideration. Experimentally, we show that our proposed method outperforms a path-ranking based algorithm and knowledge graph embedding methods on Freebase and Never-Ending Language Learning datasets.\footnote{Code and the NELL dataset are available at \url{https://github.com/xwhan/DeepPath}.}
\end{abstract}

\section{Introduction}

Deep neural networks for acoustic modeling in speech recognitionIn recent years, deep learning techniques have obtained many state-of-the-art results in various classification and recognition problems~\cite{krizhevsky2012imagenet,hinton2012deep,kim2014convolutional}. However, complex natural language processing problems often require multiple inter-related decisions, and empowering deep learning models with the ability of learning to reason is still a challenging issue.  To handle complex queries where there are no obvious answers, intelligent machines must be able to reason with existing resources, and learn to infer an unknown answer. 

More specifically, we situate our study in the context of multi-hop reasoning, which is the task of learning explicit inference formulas, given a large KG. For example, if the KG includes the beliefs such as \emph{Neymar} plays for \emph{Barcelona}, and \emph{Barcelona} are in the \emph{La Liga} league, then machines should be able to learn the following formula: 
\emph{playerPlaysForTeam(P,T)} $\land$ \emph{teamPlaysInLeague(T,L)} $\Rightarrow$ \emph{playerPlaysInLeague(P,L)}.
In the testing time, by plugging in the learned formulas, the system should be able to automatically infer the missing link between a pair of entities. This kind of reasoning machine will potentially serve as an essential components of complex QA systems.

In recent years, the Path-Ranking Algorithm (PRA)~\cite{lao2010efficient,lao2011random} emerges as a promising method for learning inference paths in large KGs. PRA uses a random-walk with restarts based inference mechanism to perform multiple bounded depth-first search processes to find relational paths. Coupled with elastic-net based learning, PRA then picks more plausible paths using supervised learning. However, PRA operates in a fully discrete space, which makes it difficult to evaluate and compare similar entities and relations in a KG. 

In this work, we propose a novel approach for controllable multi-hop reasoning: we frame the path learning process as reinforcement learning (RL). In contrast to PRA, we use translation-based knowledge based embedding method~\cite{bordes2013translating} to encode the continuous state of our RL agent, which reasons in the vector space environment of the knowledge graph. The agent takes incremental steps by sampling a relation to extend its path.
To better guide the RL agent for learning relational paths, we use policy gradient training~\cite{mnih2015human} with a novel reward function that jointly encourages accuracy, diversity, and efficiency. Empirically, we show that our method outperforms PRA and embedding based methods on a Freebase and a Never-Ending Language Learning~\cite{DBLP:conf/aaai/CarlsonBKSHM10} dataset. Our contributions are three-fold:
\begin{itemize}
\item We are the first to consider reinforcement learning (RL) methods for learning relational paths in knowledge graphs;
\item Our learning method uses a complex reward function that considers accuracy, efficiency, and path diversity simultaneously, offering better control and more flexibility in the path-finding process;
\item We show that our method can scale up to large scale knowledge graphs, outperforming PRA and KG embedding methods in two tasks.
\end{itemize}
In the next section, we outline related work in path-finding and embedding methods in KGs. We describe the proposed method in Section~\ref{sec:methodology}. We show experimental results in Section~\ref{sec:exp}. Finally, we conclude in Section~\ref{sec:conclude}.

\section{Related Work}
The Path-Ranking Algorithm (PRA) method~\cite{DBLP:conf/emnlp/LaoMC11} is a primary path-finding approach that uses random walk with restart strategies for multi-hop reasoning. Gardner et al.~\shortcite{gardner2013improving,sgardner2014incorporating} propose a modification to PRA that computes feature similarity in the vector space. Wang and Cohen~\shortcite{wang2015acl} introduce a recursive random walk approach for integrating the background KG and text---the method performs structure learning of logic programs and information extraction from text at the same time. A potential bottleneck for random walk inference is that supernodes connecting to large amount of formulas will create huge fan-out areas that significantly slow down the inference and affect the accuracy. 

Toutanova et al.~\shortcite{toutanova2015representing} provide a convolutional neural network solution to multi-hop reasoning. They build a CNN model based on lexicalized dependency paths, which suffers from the error propagation issue due to parse errors. Guu et al.~\shortcite{guu-miller-liang:2015:EMNLP} uses KG embeddings to answer path queries. Zeng et al.~\shortcite{zeng2014relation} described a CNN model for relational extraction, but it does not explicitly model the relational paths. Neelakantan et al.~\shortcite{neelakantan2015compositional} propose a recurrent neural networks model for modeling relational paths in knowledge base completion (KBC), but it trains too many separate models, and therefore it does not scale. Note that many of the recent KG reasoning methods~\cite{neelakantan2015compositional,das2016chains} still rely on first learning the PRA paths, which only operates in a discrete space. Comparing to PRA, our method reasons in a continuous space, and by incorporating various criteria in the reward function, our reinforcement learning (RL) framework has better control and more flexibility over the path-finding process. 

Neural symbolic machine~\cite{liang2016neural} is a more recent work on KG reasoning, which also applies reinforcement learning but has a different flavor from our work. NSM learns to compose programs that can find answers to natural language questions, while our RL model tries to add new facts to knowledge graph (KG) by reasoning on existing KG triples. In order to get answers, NSM learns to generate a sequence of actions that can be combined as a executable program. The action space in NSM is a set of predefined tokens. In our framework, the goal is to find reasoning paths, thus the action space is relation space in the KG. A similar framework~\cite{johnson2017inferring} has also been applied to visual reasoning tasks.

\section{Methodology}
\label{sec:methodology}
\begin{figure*}
\includegraphics[width=\textwidth]{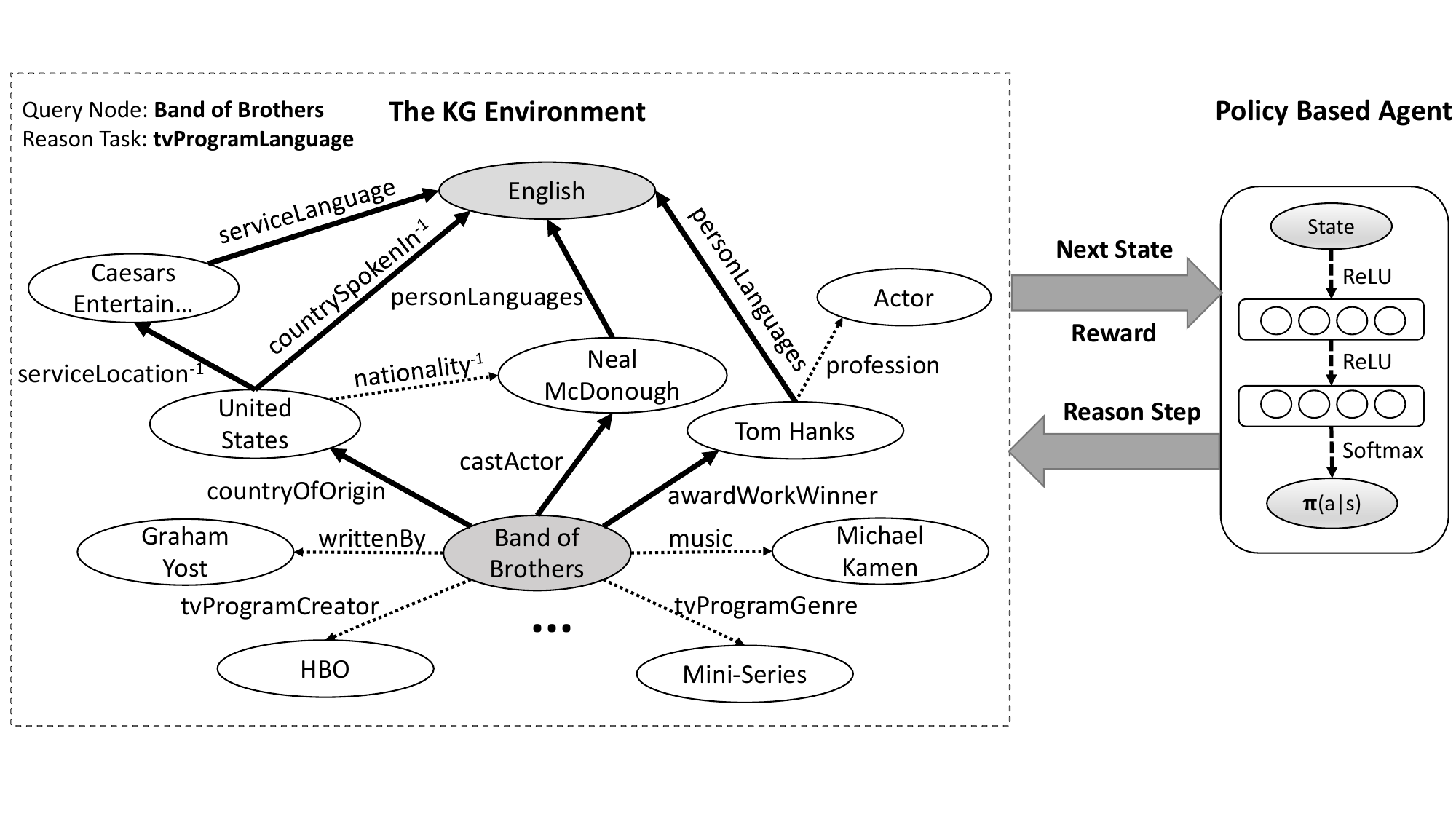}
\caption{Overview of our RL model. \textbf{Left:} The KG environment $\mathcal{E}$ modeled by a MDP. The dotted arrows (partially) show the existing relation links in the KG and the bold arrows show the reasoning paths found by the RL agent. $^{-1}$ denotes the inverse of an relation. \textbf{Right:} The structure of the policy network agent. At each step, by interacting with the environment, the agent learns to pick a relation link to extend the reasoning paths.}
\label{method}
\end{figure*}

In this section, we describe in detail our RL-based framework for multi-hop relation reasoning. The specific task of relation reasoning is to find reliable predictive paths between entity pairs. We formulate the path finding problem as a sequential decision making problem which can be solved with a RL agent. We first describe the environment and the policy-based RL agent. By interacting with the environment designed around the KG, the agent learns to pick the promising reasoning paths. Then we describe the training procedure of our RL model. After that, we describe an efficient path-constrained search algorithm for  relation reasoning with the paths found by the RL agent.
\subsection{Reinforcement Learning for Relation Reasoning}
The RL system consists of two parts (see Figure~\ref{method}). The first part is the external environment $\mathcal{E}$ which specifies the dynamics of the interaction between the agent and the KG. This environment is modeled as a Markov decision process (MDP). A tuple $<\mathcal{S,A,P,R}>$ is defined to represent the MDP, where $\mathcal{S}$ is the continuous state space, $\mathcal{A} = \{a_1,a_2,...,a_n\}$ is the set of all available actions, $\mathcal{P}(S_{t+1}=s^{'}|S_t=s,A_t=a)$ is the transition probability matrix, and $\mathcal{R}(s,a)$ is the reward function of every $(s,a)$ pairs. 

The second part of the system, the RL agent, is represented as a policy network $\pi_\theta(s,a)=p(a|s;\theta)$ which maps the state vector to a stochastic policy. The neural network parameters $\theta$ are updated using stochastic gradient descent. Compared to Deep Q Network (DQN)~\cite{Mnih}, policy-based RL methods turn out to be more appropriate for our knowledge graph scenario. One reason is that for the path finding problem in KG, the action space can be very large due to complexity of the relation graph. This can lead to poor convergence properties for DQN. Besides, instead of learning a greedy policy which is common in value-based methods like DQN, the policy network is able to learn a stochastic policy which prevent the agent from getting stuck at an intermediate state. Before we describe the structure of our policy network, we first describe the components (actions, states, rewards) of the RL environment.
\\

\noindent{\bf Actions} Given the entity pairs $(e_s,e_t)$ with relation $r$, we want the agent to find the most informative paths linking these entity pairs. Beginning with the source entity $e_s$, the agent use the policy network to pick the most promising relation to extend its path at each step until it reaches the target entity $e_t$. To keep the output dimension of the policy network consistent, the action space is defined as all the relations in the KG.
\\

\noindent{\bf States} The entities and relations in a KG are naturally discrete atomic symbols. Since existing practical KGs like Freebase~\cite{bollacker2008freebase} and NELL~\cite{carlson-aaai} often have huge amounts of triples. It is impossible to directly model all the symbolic atoms in states. To capture the semantic information of these symbols, we use translation-based embeddings such as TransE~\cite{bordes2013translating} and TransH~\cite{wang2014knowledge} to represent the entities and relations. These embeddings map all the symbols to a low-dimensional vector space. In our framework, each state captures the agent's position in the KG. After taking an action, the agent will move from one entity to another. These two are linked by the action (relation) just taken by the agent. The state vector at step $t$ is given as follows:
\begin{equation*}
\mathbf{s}_t = (\mathbf{e}_t, \mathbf{e}_{target} - \mathbf{e}_t)
\label{eq1}
\end{equation*}
where $\mathbf{e}_t$ denotes the embeddings of the current entity node and $\mathbf{e}_{target}$ denotes the embeddings of the target entity. At the initial state, $\mathbf{e}_t = \mathbf{e}_{source}$. We do not incorporate the reasoning relation in the state, because the embedding of the reasoning relation remain constant during path finding, which is not helpful in training. However, we find out that by training the RL agent using a set of positive samples for one particular relation, the agent can successfully discover the relation semantics.
\\

\noindent{\bf Rewards} There are a few factors that contribute to the quality of the paths found by the RL agent. To encourage the agent to find predictive paths, our reward functions include the following scoring criteria:

\noindent{\bf \em Global accuracy:}
For our environment settings, the number of actions that can be taken by the agent can be very large. In other words, there are much more incorrect sequential decisions than the correct ones. The number of these incorrect decision sequences can increase exponentially with the length of the path. In view of this challenge, the first reward function we add to the RL model is defined as follows:
\begin{equation*}
r_{\textsc{global}} = \begin{cases}
+1,& \text{if the path reaches } e_{target}\\ 
-1,& \text{otherwise}
\end{cases}
\end{equation*}
the agent is given an offline positive reward $+1$ if it reaches the target after a sequence of actions.

\noindent{\bf \em Path efficiency:} For the relation reasoning task, we observe that short paths tend to provide more reliable reasoning evidence than longer paths. Shorter chains of relations can also improve the efficiency of the reasoning by limiting the length of the RL's interactions with the environment. The efficiency reward is defined as follows:
\begin{equation*}
r_{\textsc{efficiency}} = \frac{1}{length(p)}
\end{equation*}
where path $p$ is defined as a sequence of relations $r_1 \rightarrow r_2 \rightarrow ... \rightarrow r_n$.

\noindent{\bf \em Path diversity:} We train the agent to find paths using positive samples for each relation. These training sample $(e_{source},e_{target})$ have similar state representations in the vector space. The agent tends to find paths with similar syntax and semantics. These paths often contains redundant information since some of them may be correlated. To encourage the agent to find diverse paths, we define a diversity reward function using the cosine similarity between the current path and the existing ones:
\begin{equation*}
r_{\textsc{diversity}} = -\frac{1}{|F|}\sum_{i=1}^{|F|}cos(\mathbf{p},\mathbf{p}_i)
\end{equation*}
where $\mathbf{p}=\sum_{i=1}^n\mathbf{r}_i$ represents the path embedding for the relation chain $r_1 \rightarrow r_2 \rightarrow ... \rightarrow r_n$. 

\noindent{\bf Policy Network} We use a fully-connected neural network to parameterize the policy function $\pi(s;\theta)$ that maps the state vector $\mathbf{s}$ to a probability distribution over all possible actions. The neural network consists of two hidden layers, each followed by a rectifier nonlinearity layer (ReLU). The output layer is normalized using a softmax function (see Figure~\ref{method}).

\subsection{Training Pipeline}
\label{subsec:train}
In practice, one big challenge of KG reasoning is that the relation set can be quite large. For a typical KG, the RL agent is often faced with hundreds (thousands) of possible actions. In other words, the output layer of the policy network often has a large dimension. Due to the complexity of the relation graph and the large action space, if we directly train the RL model by trial and errors, which is typical for RL algorithms, the RL model will show very poor convergence properties. After a long-time training, the agents fails to find any valuable path. To tackle this problem, we start our training with a supervised policy which is inspired by the  imitation learning pipeline used by {\em AlphaGo}~\cite{silver2016mastering}. In the Go game, the player is facing nearly 250 possible legal moves at each step. Directly training the agent to pick actions from the original action space can be a difficult task. {\em AlphaGo} first train a supervised policy network using experts moves. In our case, the supervised policy is trained with a randomized breadth-first search (BFS).

\noindent{\bf Supervised Policy Learning} For each relation, we use a subset of all the positive samples (entity pairs) to 
learn the supervised policy. For each positive sample $(e_{source},e_{target})$, a two-side BFS is conducted to find same correct paths between the entities. For each path $p$ with a sequence of relations $r_1 \rightarrow r_2 \rightarrow ... \rightarrow r_n$, we update the parameters $\theta$ to maximize the expected cumulative reward using Monte-Carlo Policy Gradient (REINFORCE)~\cite{williams1992simple}:
\begin{align}
J(\theta) &= \mathbb{E}_{a \sim \pi(a|s;\theta)}(\sum_{t}R_{s_t,a_t}) \nonumber \\
		  &= \sum_{t}\sum_{a\in \mathcal{A}}\pi(a|s_t;\theta)R_{s_t,a_t}
\label{eq2}
\end{align}
where $J(\theta)$ is the expected total rewards for one episode. For supervised learning, we give a reward of $+1$ for each step of a successful episode. By plugging in the paths found by the BFS, the approximated gradient used to update the policy network is shown below:
\begin{align}  
\nabla_{\theta}J(\theta) &= \sum_t\sum_{a\in \mathcal{A}}\pi(a|s_t;\theta)\nabla_{\theta}\log{\pi(a|s_t;\theta)} \nonumber \\
&\approx \nabla_{\theta}\sum_t \log{\pi(a=r_t|s_t;\theta)}
\end{align}
where $r_t$ belongs to the path $p$.

However, the vanilla BFS is a biased search algorithm which prefers short paths. When plugging in these biased paths, it becomes difficult for the agent to find longer paths which may potentially be useful. We want the paths to be controlled only by the defined reward functions. To prevent the biased search, we adopt a simple trick to add some random mechanisms to the BFS. Instead of directly searching the path between $e_{source}$ and $e_{target}$, we randomly pick a intermediate node $e_{inter}$ and then conduct two BFS between $(e_{source},e_{inter})$ and $(e_{inter},e_{target})$. The concatenated paths are used to train the agent. The supervised learning saves the agent great efforts learning from failed actions. With the learned experience, we then train the agent to find desirable paths.

\noindent{\bf Retraining with Rewards} To find the reasoning paths controlled by the reward functions, we use reward functions to retrain the supervised policy network. For each relation, the reasoning with one entity pair is treated as one episode. Starting with the source node $e_{source}$, the agent picks a relation according to the stochastic policy $\pi(a|s)$, which is a probability distribution over all relations, to extend its reasoning path. This relation link may lead to a new entity, or it may lead to nothing. These failed steps will cause the agent to receive negative rewards. The agent will stay at the same state after these failed steps. Since the agent is following a stochastic policy, the agent will not get stuck by repeating a wrong step. To improve the training efficiency, we limit the episode length with an upper bound $max\_length$. The episode ends if the agent fails to reach the target entity within $max\_length$ steps. After each episode, the policy network is updated using the following gradient:
\begin{equation}
\nabla_{\theta}J(\theta) = \nabla_{\theta}\sum_t \log{\pi(a=r_t|s_t;\theta)}R_{total}
\end{equation}
where $R_{total}$ is the linear combination of the defined reward functions. The detail of the retrain process is shown in Algorithm~\ref{retrain}. In practice, $\theta$ is updated using the Adam Optimizer~\cite{kingma2014adam} with L$_2$ regularization.

\begin{algorithm}[t]
\caption{Retraining Procedure with reward functions}\label{retrain}
Restore parameters $\theta$ from supervised policy\;
\For{episode $\leftarrow$ 1 \KwTo N}{
Initialize state vector $s_t \leftarrow s_0$\\
Initialize episode length $steps \leftarrow 0$\\
\While{$num\_steps < max\_length$}{
Randomly sample action $a\sim \pi(a|s_t)$\\
Observe reward $\mathcal{R}_t$, next state $s_{t+1}$\\
\tcp{if the step fails}
\uIf{$\mathcal{R}_t = -1$}{
Save $<s_t,a>$ to $\mathcal{M}_{neg}$
}
\uIf{success or $steps=max\_length$}{
break
}
Increment $num\_steps$
}
\tcp{penalize failed steps}
Update $\theta$ using $g \propto \nabla_{\theta}\sum_{\mathcal{M}_{neg}} \log{\pi(a=r_t|s_t;\theta)}(-1)$
\If{success}{
$R_{total} \leftarrow \lambda_1 r_{\textsc{global}} + \lambda_2 r_{\textsc{efficiency}} + \lambda_3 r_{\textsc{diversity}}$\\
Update $\theta$ using $g \propto \nabla_{\theta}\sum_{t} \log{\pi(a=r_t|s_t;\theta)}R_{total}$
}
}
\end{algorithm}

\subsection{Bi-directional Path-constrained Search}
Given an entity pair, the reasoning paths learned by the RL agent can be used as logical formulas to predict the relation link. Each formula is verified using a bi-directional search. In a typical KG, one entity node can be linked to a large number of neighbors with the same relation link. A simple example is the relation \textit{personNationality}$^{-1}$, which denotes the inverse of \textit{personNationality}. Following this link, the entity \textit{United States} can reach numerous neighboring entities. If the formula consists of such links, the number of intermediate entities can exponentially increase as we follow the reasoning formula. However, we observe that for these formulas, if we verify the formula from the inverse direction. The number of intermediate nodes can be tremendously decreased. Algorithm~\ref{search} shows a detailed description of the proposed bi-directional search.
\begin{algorithm}[t]
\caption{Bi-directional search for path verification}\label{search}
Given a reasoning path $p:{r_1 \rightarrow r_2 \rightarrow ... \rightarrow r_n}$\\
\For{$(e_i,e_j)$ in test set $\mathcal{D}$}{
start $\leftarrow$ 0; end $\leftarrow$ n\\
$left \leftarrow \emptyset;right \leftarrow \emptyset$\\
\While{start $<$ end}{
$leftEx \leftarrow \emptyset;rightEx \leftarrow \emptyset$\\
\If{len(left) $<$ len(right)}{
Extend path on the left side\\
Add connected nodes to $leftEx$\\
$left \leftarrow leftEx$
}
\Else{
Extend path on the right side\\
Add connected nodes to $rightEx$\\
$right \leftarrow rightEx$
}
}
\If{$left\cap right \neq \emptyset$ }{
return \textbf{True}
}
\Else{
return \textbf{False}}
}
\end{algorithm}

\section{Experiments}
\label{sec:exp}
To evaluate the reasoning formulas found by our RL agent, we explore two standard KG reasoning tasks:  link prediction (predicting target entities) and fact prediction (predicting whether an unknown fact holds or not). We compare our method with both path-based methods and embedding based methods. After that, we further analyze the reasoning paths found by our RL agent. These highly predictive paths validate the effectiveness of the reward functions. Finally, we conduct a experiment to investigate the effect of the supervised learning procedure.
\subsection{Dataset and Settings}
\begin{table}[h!]
\small
\centering
 \begin{tabular}{||c c c c c||} 
 \hline
 Dataset & \# Ent. & \# R. & \# Triples & \# Tasks\\ 
 \hline\hline
FB15K-237 & 14,505 & 237 & 310,116 & 20 \\ 
 NELL-995 & 75,492 & 200 & 154.213 & 12\\
 \hline
 \end{tabular}
 \caption{Statistics of the Datasets. \# Ent. denotes the number of unique entities and \# R. denotes the number of relations}
 \label{stats}
\end{table}

\begin{table*}[h]
\centering
\small
\begin{tabular}{ccccccccccc}\toprule
& \multicolumn{4}{c}{FB15K-237} & & \multicolumn{4}{c}{NELL-995} \\
\cmidrule{2-5} \cmidrule{7-10}
Tasks & PRA & RL & TransE & TransR & Tasks & PRA & RL & TransE & TransR \\ \midrule
teamSports & \textbf{0.987} & 0.955 & 0.896 & 0.784& athletePlaysForTeam & 0.547 &\textbf{0.750} & 0.627 & 0.673\\
birthPlace & 0.441 & \textbf{0.531} & 0.403 & 0.417 & athletePlaysInLeague & 0.841 & \textbf{0.960} & 0.773 & 0.912\\
personNationality & \textbf{0.846} & 0.823 & 0.641 & 0.720 & athleteHomeStadium & 0.859 & \textbf{0.890} & 0.718 & 0.722\\
filmDirector & 0.349 & \textbf{0.441} & 0.386 & 0.399 & athletePlaysSport& 0.474& 0.957 & 0.876 & \textbf{0.963}\\
filmWrittenBy & 0.601 & 0.457 & 0.563 & \textbf{0.605} & teamPlaySports& 0.791& 0.738 & 0.761 & \textbf{0.814}\\
filmLanguage & 0.663 & \textbf{0.670} & 0.642 & 0.641 & orgHeadquaterCity& \textbf{0.811}& 0.790 & 0.620 & 0.657 \\ 
tvLanguage & 0.960 & \textbf{0.969} & 0.804 & 0.906 & worksFor & 0.681 & \textbf{0.711} & 0.677 & 0.692 \\
capitalOf & \textbf{0.829} & 0.783 & 0.554 & 0.493 &bornLocation & 0.668 & 0.757 & 0.712 & \textbf{0.812}\\
organizationFounded & 0.281 & 0.309 & \textbf{0.390} & 0.339 & personLeadsOrg & 0.700 &\textbf{0.795} & 0.751 & 0.772\\
musicianOrigin & 0.426 & \textbf{0.514} & 0.361 & 0.379 & orgHiredPerson & 0.599 & \textbf{0.742} & 0.719 & 0.737 \\
... & &  & & & ...\\
\midrule
Overall & 0.541 & \textbf{0.572} & 0.532 & 0.540 &  & 0.675& \textbf{0.796} & 0.737 & 0.789\\
\bottomrule
\end{tabular}
\caption{Link prediction results (MAP) on two datasets.}
\label{result1}
\end{table*}

Table~\ref{stats} shows the statistics of the two datasets we conduct our experiments on. Both of them are subsets of larger datasets. The triples in FB15K-237~\cite{toutanova2015representing} are sampled from FB15K~\cite{bordes2013translating} with redundant relations removed. We perform the reasoning tasks on 20 relations which have enough reasoning paths. These tasks consists of relations from different domains like \textit{Sports}, \textit{People}, \textit{Locations}, \textit{Film}, etc. Besides, we present a new NELL subset that is suitable for multi-hop reasoning from the 995th iteration of the NELL system. We first remove the triples with relation \textit{generalizations} or \textit{haswikipediaurl}. These two relations appear more than 2M times in the NELL dataset, but they have no reasoning values. After this step, we only select the triples with Top-200 relations. To facilitate path finding, we also add the inverse triples. For each triple $(h,r,t)$, we append $(t,r^{-1},h)$ to the datasets. With these inverse triples, the agent is able to step backward in the KG.

For each reasoning task $r_{i}$, we remove all the triples with $r_i$ or $r_i^{-1}$ from the KG. These removed triples are split into train and test samples. For the link prediction task, each $h$ in the test triples $\{(h,r,t)\}$ is considered as one query. A set of candidate target entities are ranked using different methods. For fact prediction, the true test triples are ranked with some generated false triples.

\subsection{Baselines and Implementation Details}
Most KG reasoning methods are based on either path formulas or KG embeddings. we explore methods from both of these two classes in our experiments. For path based methods, we compare our RL model with the PRA~\cite{lao2011random} algorithm, which has been used in a couple of reasoning methods~\cite{gardner2013improving,neelakantan2015compositional}. PRA is a data-driven algorithm using random walks (RW) to find paths and obtain path features. For embedding based methods, we evaluate several state-of-the-art embeddings designed for knowledge base completion, such as  TransE~\cite{bordes2013translating}, TransH~\cite{wang2014knowledge}, TransR~\cite{lin2015learning} and TransD~\cite{ji2015knowledge} .

The implementation of PRA is based on the code released by~\cite{lao2011random}. We use the TopK negative mode to generate negative samples for both train and test samples. For each positive samples, there are approximately 10 corresponding negative samples. Each negative sample is generated by replacing the true target entity $t$ with a faked one $t^{'}$ in each triple $(h,r,t)$. These positive and negative test pairs generated by PRA make up the test set for all methods evaluated in this paper. For Trans{E,R,H,D}, we learn a separate embedding matrix for each reasoning task using the positive training entity pairs. All these embeddings are trained for 1,000 epochs. \footnote{The implementation we used can be found at \url{https://github.com/thunlp/Fast-TransX}}

Our RL model make use of TransE to get the continuous representation of the entities and relations. We use the same dimension as Trans{E, R} to embed the entities. Specifically, the state vector we use has a dimension of 200, which is also the input size of the policy network. To reason using the path formulas, we adopt a similar linear regression approach as in PRA to re-rank the paths. However, instead of using the random walk probabilities as path features, which can be computationally expensive, we simply use binary path features obtained by the bi-directional search. We observe that with only a few mined path formulas, our method can achieve better results than PRA's data-driven approach.
\subsection{Results}

\begin{table}[t]
\centering
\begin{tabular}{cc|c}
\toprule
& \multicolumn{2}{c}{Fact Prediction Results}\\
\cmidrule{2-3}
Methods & FB15K-237 & NELL-995 \\ \midrule
RL & \textbf{0.311} & \textbf{0.493}\\
TransE & 0.277 & 0.383\\
TransH & 0.309 & 0.389\\
TransR & 0.302 & 0.406\\
TransD & 0.303 & 0.413\\
\bottomrule
\end{tabular}
\caption{Fact prediction results (MAP) on two datasets.}
\label{result2}
\end{table}

\subsubsection{Quantitative Results}
\begin{table}[t]
\centering
\begin{tabular}{ccc}
\toprule
& \multicolumn{2}{c}{\# of Reasoning Paths}\\
\cmidrule{2-3}
Tasks & PRA & RL \\ \midrule
worksFor & 247 & 25\\
teamPlaySports & 113 & 27\\
teamPlaysInLeague & 69 & 21\\
athletehomestadium & 37 & 11\\
organizationHiredPerson & 244 & 9\\
... & &\\
Average \# &137.2 & 20.3\\
\bottomrule
\end{tabular}
\caption{Number of reasoning paths used by PRA and our RL model. \emph{RL achieved better MAP with a more compact set of learned paths.}}
\label{result3}
\end{table}
\noindent{\bf Link Prediction} This task is to rank the target entities given a query entity. Table~\ref{result1} shows the mean average precision (MAP) results on two datasets. Since path-based methods generally work better than embedding methods for this task, we do not include the other two embedding baselines in this table. Instead, we spare the room to show the detailed results on each relation reasoning task.

For the overall MAP shown in the last row of the table, our approach significantly outperforms both the path-based method and embedding methods on two datasets, which validates the strong reasoning ability of our RL model. For most relations, since the embedding methods fail to use the path information in the KG, they generally perform worse than our RL model or PRA. However, when there are not enough paths between entities, our model and PRA can give poor results. For example, for the relation \textit{filmWrittenBy}, our RL model only finds 4 unique reasoning paths, which means there is actually not enough reasoning evidence existing in the KG. Another observation is that we always get better performance on the NELL dataset. By analyzing the paths found from the KGs, we believe the potential reason is that the NELL dataset has more short paths than FB15K-237 and some of them are simply synonyms of the reasoning relations.

\noindent{\bf Fact Prediction} Instead of ranking the target entities, this task directly ranks all the positive and negative samples for a particular relation. The PRA is not included as a baseline here, since the PRA code only gives a target entity ranking for each query node instead of a ranking of all triples. Table~\ref{result2} shows the overall results of all the methods. Our RL model gets even better results on this task. We also observe that the RL model beats all the embedding baselines on most reasoning tasks.

\begin{table*}[t]
\centering
\begin{tabular}{ll}\toprule
\multicolumn{1}{l}{Relation} & \multicolumn{1}{l}{Reasoning Path}\\ \midrule
\multirow{3}*{\textbf{filmCountry}} & filmReleaseRegion\\
& featureFilmLocation $\rightarrow$ locationContains$^{-1}$\\
& actorFilm$^{-1} \rightarrow$ personNationality\\ \midrule
\multirow{3}*{\textbf{personNationality}} & placeOfBirth$ \rightarrow$ locationContains$^{-1}$\\
& peoplePlaceLived $\rightarrow$ locationContains$^{-1}$\\
& peopleMarriage $\rightarrow$ locationOfCeremony $\rightarrow$ locationContains$^{-1}$\\ \midrule
\multirow{3}*{\textbf{tvProgramLanguage}} & tvCountryOfOrigin $\rightarrow$ countryOfficialLanguage \\
& tvCountryOfOrigin $\rightarrow$ filmReleaseRegion$^{-1} \rightarrow$ filmLanguage\\
& tvCastActor $\rightarrow$ filmLanguage\\ \midrule 
\multirow{3}*{\textbf{personBornInLocation}} & personBornInCity\\
& graduatedUniversity $\rightarrow$ graduatedSchool$^{-1} \rightarrow$ personBornInCity \\
& personBornInCity $\rightarrow$ atLocation$^{-1} \rightarrow$ atLocation\\ \midrule
\multirow{3}*{\textbf{athletePlaysForTeam}} 
& athleteHomeStadium $\rightarrow$ teamHomeStadium$^{-1}$ \\
& athletePlaysSport $\rightarrow$ teamPlaysSport$^{-1}$\\
& athleteLedSportsTeam\\ \midrule
\multirow{3}*{\textbf{personLeadsOrganization}}
& worksFor\\
& organizationTerminatedPerson$^{-1}$\\
& mutualProxyFor$^{-1}$\\
\bottomrule
\end{tabular}
\caption{Example reasoning paths found by our RL model. The first three relations come from the FB15K-237 dataset. The others are from NELL-995. Inverses of existing relations are denoted by $^{-1}$. }
\label{formulas}
\end{table*}

\subsubsection{Qualitative Analysis of Reasoning Paths}

\begin{figure}[t]
\centering
\includegraphics[width=0.9\linewidth]{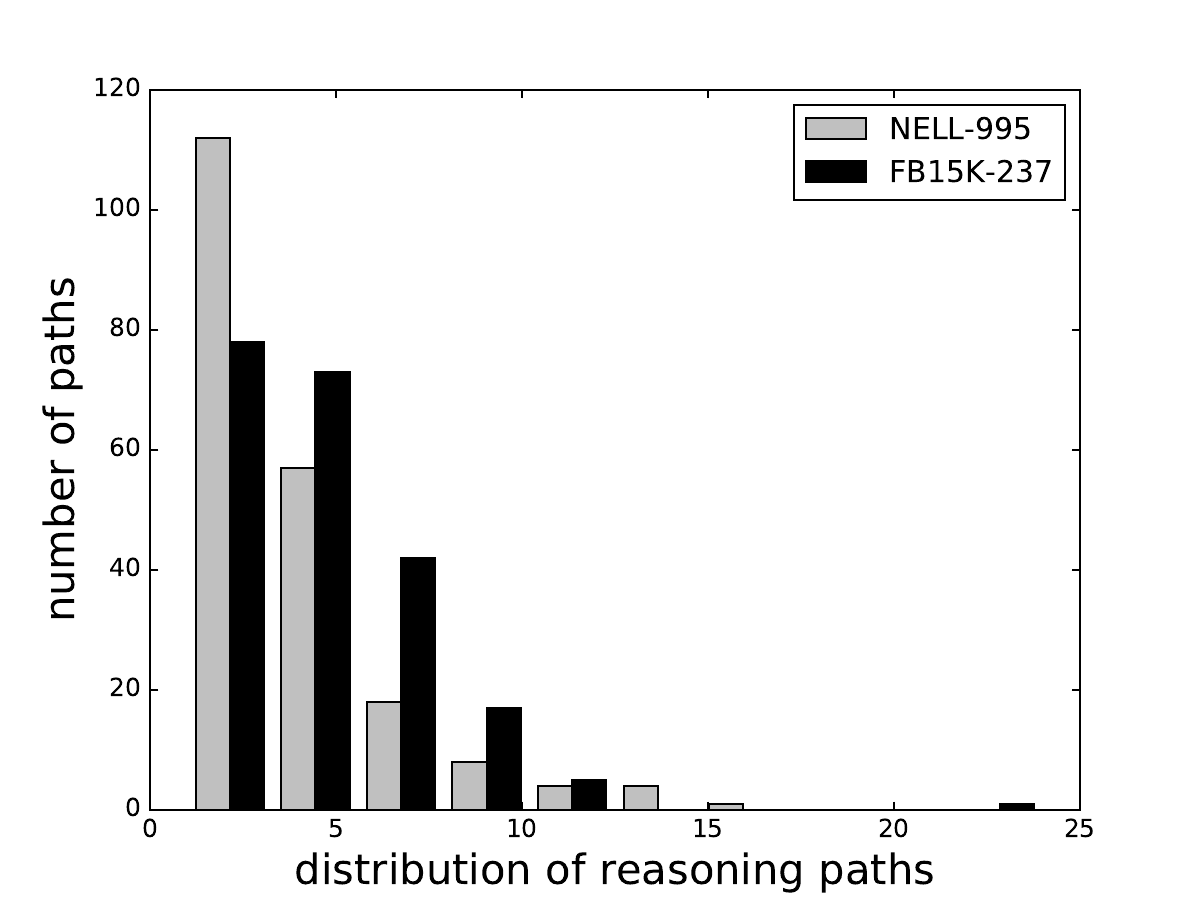}
\caption{The distribution of paths lengths on two datasets}
\label{hist}
\end{figure}

To analyze the properties of reasoning paths, we show a few reasoning paths found by the agent in Table~\ref{formulas}. To illustrate the effect of the efficiency reward function, we show the path length distributions in Figure~\ref{hist}. To interpret these paths, take the \textit{personNationality} relation for example, the first reasoning path indicates that if we know facts \textit{placeOfBirth(x,y)} and \textit{locationContains(z,y)} then it is highly possible that person $x$ has nationality $z$. These short but predictive paths indicate the effectiveness of the RL model. Another important observation is that our model use much fewer reasoning paths than PRA, which indicates that our model can actually extract the most reliable reasoning evidence from KG. Table~\ref{result3} shows some comparisons about the number of reasoning paths. We can see that, with the pre-defined reward functions, the RL agent is capable of picking the strong ones and filter out similar or irrelevant ones.

\begin{figure}[t]
\centering
\includegraphics[width=0.98\linewidth]{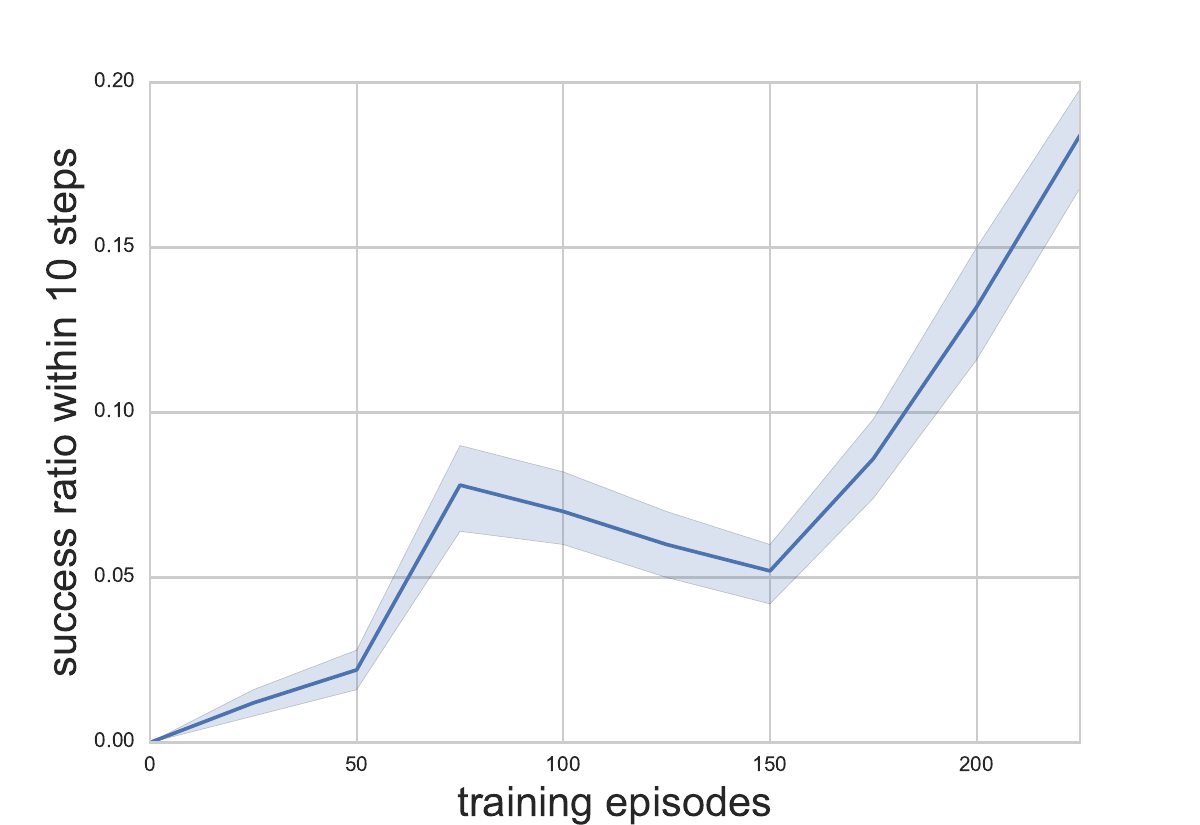}
\caption{The success ratio ($succ_{10}$) during training. Task: athletePlaysForTeam.\footnotemark}
\label{stats1}
\end{figure}

\footnotetext{The confidence band is generated using 50 different runs.}

\subsubsection{Effect of Supervised Learning}
As mentioned in Section~\ref{subsec:train}, one major challenge for applying RL to KG reasoning is the large action space. We address this issue by applying supervised learning before the reward retraining step. To show the effect of the supervised training, we evaluate the agent's success ratio of reaching the target within 10 steps ($succ_{10}$) after different number of training episodes. For each training episode, one pair of entities $(e_{source},e_{target})$ in the train set is used to find paths. All the correct paths linking the entities will get a $+1$ global reward. We then plug in some true paths for training. The $succ_{10}$ is calculated on a held-out test set that consists of 100 entity pairs. For the NELL-995 dataset, since we have 200 unique relations, the dimension of the action space will be 400 after we add the backward actions. This means that random walks will get very low $succ_{10}$ since there may be nearly $400^{10}$ invalid paths. Figure~\ref{stats1} shows the $succ_{10}$ during training. We see that even the agent has not seen the entity before, it can actually pick the promising relation to extend its path. This also validates the effectiveness of our state representations.

\section{Conclusion and Future Work}
\label{sec:conclude}
In this paper, we propose a reinforcement learning framework to improve the performance of relation reasoning in KGs. Specifically, we train a RL agent to find reasoning paths in the knowledge base. Unlike previous path finding models that are based on random walks, the RL model allows us to control the properties of the found paths. These effective paths can also be used as an alternative to PRA in many path-based reasoning methods. For two standard reasoning tasks, using the RL paths as reasoning formulas, our approach generally outperforms two classes of baselines. 

For future studies, we plan to investigate the possibility of incorporating adversarial learning~\cite{goodfellow2014generative} to give better rewards than the human-defined reward functions used in this work. Instead of designing rewards according to path characteristics, a discriminative model can be trained to give rewards. Also, to address the problematic scenario when the KG does not have enough reasoning paths, we are interested in applying our RL framework to joint reasoning with KG triples and text mentions.

\section*{Acknowledgments}
We gratefully acknowledge the support of NVIDIA Corporation with the donation of one Titan X Pascal GPU used for this research.

\bibliography{emnlp2017}
\bibliographystyle{emnlp_natbib}

\end{document}